# The Integration of Connectionism and First-Order Knowledge Representation and Reasoning as a Challenge for Artificial Intelligence

Sebastian Bader[1], Pascal Hitzler[2], Steffen Hölldobler[1]

[1]International Center for Computational Logic
Technische Universität Dresden, Germany
[2]AIFB, Universität Karlsruhe, Germany

**Abstract**

Intelligent systems based on first-order logic on the one hand, and on artificial neural networks (also called connectionist systems) on the other, differ substantially. It would be very desirable to combine the robust neural networking machinery with symbolic knowledge representation and reasoning paradigms like logic programming in such a way that the strengths of either paradigm will be retained. Current state-of-the-art research, however, fails by far to achieve this ultimate goal. As one of the main obstacles to be overcome we perceive the question how symbolic knowledge can be encoded by means of connectionist systems: Satisfactory answers to this will naturally lead the way to knowledge extraction algorithms and to integrated neural-symbolic systems.

## 1 Introduction

Artificial neural networks — also called *connectionist systems* — exhibit many desirable properties of intelligent systems like, for example, being massively parallel, context-sensitive, adaptable and robust (see eg. [14]). It is strongly believed that intelligent systems must also be able to represent and reason about structured objects and structure-sensitive processes (see eg. [16, 35]). Unfortunately, we are unaware of any connectionist system which can handle structured objects and structure-sensitive processes in a satisfying way. Logic systems were designed to cope with such objects and processes and, consequently, it is a long-standing research goal to combine the advantages of connectionist and logic systems in a single system.

Earlier attempts to integrate logic and connectionist systems have mainly been restricted to propositional logic, or to first-order logic without function symbols. They go back to the pioneering work by McCulloch and Pitts [34], and have led to a number of systems developed in the 80s and 90s, including Towell and Shavlik's KBANN [45], Shastri and Ajjanagadde's SHRUTI [43], Lange and Dryer's ROBIN [32] the work by Pinkas [37], Hölldobler [22], and d'Avila Garcez et al. [10, 13], to mention a few, and we refer to [9, 11] for comprehensive literature overviews.

Without the restriction to the finite case (including propositional logic and first-order logic without function symbols), the task becomes much harder due to the fact that the underlying language is infinite but shall be encoded using networks with a finite number of nodes. One of the few approaches for overcoming this problem (apart from work on recursive autoassociative memory, RAAM, initiated by Pollack [40], which concerns the learning of recursive terms over a first-order language) is based on a proposal by Hölldobler et al. [27], and reported also in [18]. It is based on the idea that logic programs can be represented by their associated single-step or immediate consequence operators. Such an operator can then be mapped to a function on the real numbers, which can under certain conditions in turn be encoded or approximated e.g. by feedforward networks with sigmoidal activation functions.

The purpose of this paper is twofold. First, we will give an overview of recent progress made in the representation of first-order logic programs by connectionist systems (Section 2). We will then discuss in detail some questions which we find of central importance in order to advance towards an integration of logic and connectionism (Section 3). Our selections are certainly very subjective, so we also provide ample references to related work and literature for further reading. Our discussions will be in very general terms, and we will make most of our general exhibition accessible to the general reader. Some familiarity with basic notions from logic and artificial neural networks, and also from set-theoretic topology and iterated function systems will however be helpful for understanding some of the details. As general references we recommend [33, 7, 47, 4].

*Acknowledgements.* The first author is supported by the GK334 of the German Research Foundation. The second author is supported by the German Federal Ministry of Education and Research under the SmartWeb project and the EU Network of Excellence KnowledgeWeb. The second and third author acknowledge substantial support by the Boole Centre for Research in Informatics at the National University of Ireland, Cork, for presenting this paper.

## 2   Recent Progress

Integrating first-order logical knowledge representation and connectionism necessitates to find a common framework in which both kinds of systems can be expressed and somehow unified.

Logical knowledge representation is symbolic in nature, i.e. the data structures under consideration basically consist of words over some language or of collections of finite trees, for example, depending on the perspective taken or on the problem at hand. Logic programs, more specifically, consist of sets of first-order formulae under a restricted syntax, more precisely, a logic program is a set of (universally quantified) disjunctions, called *clauses* or *rules*, which in turn consist of atoms and negated atoms only. Equivalently, one can say that logic programs are basically formulae in conjunctive normal form — although their meaning, i.e. the way they are evaluated, is not identical to their meaning in first-order logic. Input (queries) and output (answers) of a logic program essentially consist of certain logical formulae or of models of the program.

Successful connectionist architectures, however, can be understood as networks (essentially, directed graphs) of simple computational units, in which activation is propagated and combined in certain ways adhering to connectionist principles. In many cases like, for example, in multilayer perceptrons, the activation is encoded as a real number; input and output of such networks consist of tuples (vectors) of real numbers. So, while logic is *symbolic* and thus *discrete*, standard connectionist systems are *continuous*, i.e. they deal with real values in Euclidean space.

In order to integrate logic and connectionism we thus need to bridge the gap between the discrete, symbolic setting of logic, and the continuous, real-valued setting of artificial neural networks. The method of our choice — motivated by several reasons which will become clear below — is to employ *Cantor space* for this purpose.

Cantor space $\mathcal{C}$ is — up to homeomorphism — a subset of the unit interval of the real numbers endowed with the topological structure inherited from the reals. The set is best described as the set of all real numbers in the unit interval which can be expressed in the ternary number system using the digits 0 and 2 only. More precisely, $\mathcal{C}$ is the set of all real numbers of the form $\sum_{i=1}^{\infty} a_i 3^{-i}$, where $a_i \in \{0, 2\}$ for all $i$. We remark that topologically, we obtain homeomorphic subsets of the reals by considering all real numbers of the form $\sum_{i=1}^{\infty} a_i B^{-i}$, where $a_i \in \{0, 1\}$ and $B$ is fixed to some natural number greater than or equal to 3. This lies in the fact that Cantor space can be uniquely described — up to homeomorphism — as the topological space which is totally disconnected, compact, Hausdorff, second countable, and dense in itself.

How do we relate Cantor space to first-order logic? The topological characterization of $\mathcal{C}$ just given already shows that it can be described independently of the real numbers. Now consider some first-order language $\mathcal{L}$. Interpretations (or valuations) over $\mathcal{L}$ can be understood as mappings from the countable set of ground atoms over $\mathcal{L}$ — which we call the *Herbrand base* $B_\mathcal{L}$ over $\mathcal{L}$ — to the set of truth values $\{\mathbf{t}, \mathbf{f}\}$. Identifying $\mathbf{t}$ with 2 (or 1) and $\mathbf{f}$ with 0, the set of all interpretations over $\mathcal{L}$ can be identified with the set of all mappings from $B_\mathcal{L}$ to $\{0, 2\}$. Since $B_\mathcal{L}$ is countable, we can also choose an enumeration of $B_\mathcal{L}$, which is essentially an identification of $B_\mathcal{L}$ with $\mathbb{N}$, the set of natural numbers excluding zero, or, in other words, a bijective mapping $l : B_\mathcal{L} \to \mathbb{N}$. We can thus identify the set of all interpretations over $\mathcal{L}$, which are of the form $I : B_\mathcal{L} \to \{\mathbf{t}, \mathbf{f}\}$, with the set of all mappings $f : \mathbb{N} \to \{0, 2\}$.

Now, formally, let $l : B_\mathcal{L} \to \mathbb{N}$ be an (arbitrarily chosen) bijection and let $I_\mathcal{L}$ be the set of all interpretations over $\mathcal{L}$, i.e. the set of all mappings from $B_\mathcal{L}$ to $\{0, 2\}$. We define a mapping $\iota$ from $I_\mathcal{L}$ to $\mathcal{C}$ by

$$\iota(I) = \sum_{i=1}^{\infty} I(l^{-1}(i)) 3^{-i}.$$

It is easily verified that $\iota$ is a bijection.

The mapping $\iota$ allows to understand the set of interpretations as the Cantor set in the real line. But does it also preserve meaningful structure, i.e. does it relate meaningful structure for logic programs on the one side with meaningful structure for connectionist sytems on the other side? We will see that it does, and in order to proceed, we reproduce next a theorem due to Funahashi [17].

**Theorem 1** *Suppose that $\phi : \mathbb{R} \to \mathbb{R}$ is non-constant, bounded, monotone increasing and continuous. Let $K \subseteq \mathbb{R}^n$ be compact, let $f : K \to \mathbb{R}$ be a continuous mapping and let $\varepsilon > 0$. Then there exists a 3-layer feedforward network with squashing function $\phi$ whose input-output mapping $\bar{f} : K \to \mathbb{R}$ satisfies $\max_{x \in K} d(f(x), \bar{f}(x)) < \varepsilon$, where $d$ is a metric which induces the natural topology on $\mathbb{R}$.*

For the reader who is not familiar with the terminology of the theorem, we state its intuitive meaning: Every continuous real-valued function defined on a compact subset of the reals can be uniformly approximated by input-output mappings of artificial neural networks of a certain architecture. The details of this architecture will not concern us for our general discussion.

Funahashi's theorem provides an existence result for approximating continous functions on the reals. So if we manage to interpret logic programs as such functions in a meaningful way, then we know that approximation of logic programs by neural networks is possible in a reasonable way. We need two more steps in order to realize this idea.

Firstly, we note that it is very common in logic programming to associate operators to logic programs in such a way that the behaviour of the operator reflects the meaning of the program. One of the most popular — and arguably the most natural — operator is the so-called *immediate consequence operator* $T_P$ associated with a given program $P$. Details of the definition of $T_P$ will be of no concern for our general discussion, so we will not spell them out. For the same reason, we also omit a formal definition of a logic program, and just remark that logic programs are certain sets of first-order logical formulae, as already mentioned. The operator $T_P$ is an operator which acts on $I_\mathcal{L}$, i.e. on the space of all interpretations of the underlying first-order language. Since $\iota$ maps $I_\mathcal{L}$ bijectively onto $\mathcal{C}$, we can carry over the operator $T_P$ via $\iota$ to the reals, by defining
$$\iota(T_P) : \mathcal{C} \to \mathcal{C} : x \mapsto \iota(T_P(\iota^{-1}(x))).$$

Hence, $\iota(T_P)$ is a mapping on Cantor set which carries the meaning of $P$.

Secondly, we need to ensure that the embedded mapping $\iota(T_P)$ just defined is continuous on Cantor space, such that Funahashi's theorem can be applied. This, for example, is the case if $T_P$ is continuous with respect to the initial topology induced by $\iota$ on $I_\mathcal{L}$ — let us denote this topology by $Q$. Since $\iota$ is a bijection, it follows that it is a homeomorphism from $(I_\mathcal{L}, Q)$ to Cantor space $\mathcal{C}$ — i.e. $(I_\mathcal{L}, Q)$ *is* Cantor space up to homeomorphism, and $\iota(T_P)$ is continuous as a function on $\mathcal{C}$ if and only if $T_P$ is continuous as a function on $(I_\mathcal{L}, Q)$. Together, we obtain the following result, which was reported in [18] in a more general form.

**Theorem 2** *Let $P$ be a logic program such that $T_P$ is continuous in $Q$, and let $\iota$ be a homeomorphism from $(I_\mathcal{L}, Q)$ to $\mathcal{C}$. Then $\iota(T_P)$ can be approximated uniformly by input-output functions of artificial neural networks of the kind used in Theorem 1.*

The importance of Theorem 2 lies in the fact that the topology $Q$ on $I_\mathcal{L}$ is well-known in logic programming. Indeed, it is the most important topology for the study of fixed-point semantics for programs with negation. It dates back to the work by Batarekh and Subrahmanian [5] where it was called the *query topology*. Seda [42] studies a generalization of it under the name *atomic topology*, and in the same paper it was also shown that continuity in $Q$ can naturally be characterized without making reference to topological notions. It is also strongly related to the studies of fixed-point semantics of logic programs by means of generalized metrics, as e.g. undertaken by Fitting [15], Prieß-Crampe & Ribenboim [41] and Seda & Hitzler [20].

Due to its very general nature, Theorem 2 carries a lot of inherent flexibility. The particular instance of $\iota$ given earlier is only one very specific example of a homeomorphism which can be used. Indeed, the number of automorphisms of Cantor space is uncountable. The specific representations of Cantor space as a subspace of the reals given earlier are also just very particular examples of such representations. Results analogous to that by Funahashi furthermore hold for many popular neural network architectures, such that our investigations are not a priori restricted to certain types of connectionist systems.

But the flexibility gained by the general nature of Theorem 2 does not come for free. In particular, it provides no means of actually obtaining an approximating network from a concretely given program. At best, we would like to be able to read off parameters for an approximating network directly from a given program. To date, it is an open problem how to do this along the lines of Theorem 2.

A different approach towards obtaining concrete approximations was undertaken by us in [2]. It was based on the observation that graphs of embedded operators $\iota(T_P)$, displayed in the Euclidean plane, exhibit self-similar structures known from chaos theory. More precisely, the graphs appeared to be attractors of iterated function systems as studied, for example, in the well-known book by Barnsley [4]. This led to the following theorem, which is stated in slightly more general form in [2].

**Theorem 3** *Let $P$ be a logic program such that $\iota(T_P)$ is Lipschitz-continuous. Then there exists an iterated function system on the Euclidean plane whose attractor is the graph of $\iota(T_P)$.*

The importance of Theorem 3 lies in the fact that iterated function systems can be encoded very easily as some standard type of recurrent neural networks, and we have spelled this out in [2]. The very general Theorem 3 also leads to concrete instances of iterated function systems — and thus of corresponding networks — for approximating $\iota(T_P)$: Given a program $P$ and an arbitrarily chosen accuracy of the approximation $i \in \mathbb{N}$, we need only determine a finite number of explicitly determined function values of $\iota(T_P)$, in order to arrive at an iterated function system $\mathcal{S}_i$ whose attractor $f_i$ is the graph of a continuous function — details of the construction can be found in [2]. Our result now reads as follows.

**Theorem 4** *Let $P$ be a program with Lipschitz-continuous $\iota(T_P)$. Then the sequence $(f_i)_{i \in \mathbb{N}}$ of attractors, as mentioned above, converges uniformly to $\iota(T_P)$.*

A concrete open problem remaining with Theorem 4 is that the determination of a suitable iterated function system $\mathcal{S}_i$ from a given program $P$ hinges on the explicit knowledge about an upper bound for the Lipschitz-constant for $\iota(T_P)$ — if it exists at all.

We close our brief survey with a number of further remarks.

(1) The idea to represent a logic program via its semantic operator traces back to Hölldobler & Kalinke [23], surveyed in [18], where this idea was employed for the propositional case. D'Avila Garcez et al. [10, 13] have molded this into an integrated learning system which uses backpropagation.

(2) A very restricted version of Theorem 2 was shown in [27] using different methods. There, and in [18], the network architecture was also extended in order to mimic iterations of the immediate consequence operator, and corresponding results on the convergence behaviour and speed of these iterations were provided.

(3) It was shown in [18] that many semantic operators in logic programming, including the immediate consequence operator, are measurable. While there exist approximation results relating measurability to artificial neural networks, e.g. by Hornik et al. [29], it is an open issue whether this fact can be exploited for neural-symbolic integration.

(4) A recent result by Wendt [46] relates semantic operators used in answer set programming [44] to the immediate consequence operator, and thus allows to use our results for studying non-monotonic reasoning with logic programs in a connectionist setting. This remains to be spelled out. Some preliminary investigations can be found in [19].

(5) We are recently investigating the use of weighted automata and fibring neural networks for our purposes [3, 12].

After this survey on the current state of the art of relating logic programs and connectionist networks we will identify a number of open research problems in the following section.

## 3  Challenges

### 3.1  How can first-order terms be represented and manipulated in a connectionist system?

This is the main question that needs to be answered, and our recent results presented in the previous section are along this line. We consider this question to be of central importance because the development of a satisfactory and usable representation of first-order formulae is the first necessary step towards neural-symbolic integration. The proposals made so far do not give a satisfying answer to this question: Structured connectionist networks as used e.g. in [21] are completely local. The unification and matching operations can directly be implemented in these networks. However, the number of units is quadratic and the number of connections even cubic wrt the size of the terms. It is not obvious at all how such networks can be learned.

Vectors of fixed length are used to represent terms in the recursive auto-associative memory and its derivatives [39, 1]. Unfortunately, in extensive tests none of these proposals has led to satisfying results: The systems could not safely store and recall terms of depth larger than five [30].

In hybrid systems terms are represented and manipulated in a conventional way. But this is not a kind of integration we are hoping for because in this case results from connectionist systems cannot be applied to the conventional part.

The phase-coding mechanism suggested in SHRUTI [43] and used in the BUR system [25] restricts the first-order language to contain only constants and multi-place relation symbols.

We definitely need new ideas to solve this challenge problem! Connectionist encodings for conventional data structures like counters and stacks [24, 31] have been proposed and may be of use, and the study of relationships between logic programs, neural networks, and other paradigms in computing and mathematics like cellular and weighted automata, dynamical systems, and the like, may provide new ideas.

## 3.2 How can first-order rules be extracted from a connectionist network?

To the best of our knowledge all rule extraction techniques for connectionist networks are propositional in the sense that they only generate propositional rules. For example, the propositional networks in [23] were slightly modified in [13, 10] such that backpropagation could be applied and standard rule extraction techniques could be used to extract new revised — but propositional — rules.

The results from [18] guarantee the existence of recurrent networks with a feed forward kernel to approximate the meaning of a first-order program. Backpropagation can again in principle be used to train these kernels. But it is by no means obvious how the rule extraction techniques known so far can be generalized such that first-order rules are extracted from these kernels.

## 3.3 How can distributed knowledge representation in connectionist networks be understood from a symbolic perspective?

Although this question is being subsumed by the previous two, we want to emphasize the difficulties underlying distributed representations explicitly. The representation of propositional logic in connectionist networks most often is very local, while standard network training normally leads to distributed representation, which is very difficult to interpret in a symbolic manner.

The situation becomes worse for first-order logic, where due to the infinitary nature of the underlying language there seems to be no way at all to avoid distributed representation. We understand that this issue provides the main obstacle in developing constructive methods for the representation of first-order logic programs by means of Funahashi's theorem, and we also expect this to be a major issue in order to make advances in first-order rule extraction.

## 3.4 How can established learning algorithms like backpropagation be combined with symbolic knowledge representation?

For the propositional system developed by d'Avila Garcez et al. [10, 13], symbolic knowledge is being represented by a network, which is then trained using backpropagation. Afterwards, the learned knowledge may be extracted. A similiar approach underlies the KBANN system due to Towell and Shavlik [45].

While this is a good idea, we see the risk that the initial knowledge may be lost in the course of the training process, although it should rather influence it. We envision an integration via continuous interaction of standard learning techniques with background and dynamically acquired knowledge. How this can be achieved, however, is as yet entirely unclear.

Another problem is posed by the fact that the representation of first-order knowledge easily leads to non-standard network architectures, like in the SHRUTI system [43], which cannot be trained easily, or at least cannot be trained with established methods without loosing the specific logically meaningful architecture. The latter would be the case e.g. with the recurrent networks obtained from iterated function systems as mentioned in Section 2. Modified learning algorithms will have to be established and studied for these purposes.

### 3.5 How can multiple instances of first-order rules be represented in a connectionist system?

One of the properties of first-order reasoning is that it cannot be determined in advance how many copies of a rule are needed to answer a given query or, equivalently, to prove a theorem. In local connectionist systems like CHCL [26] this problem is defined away by simply assuming that each rule is used only once. A similar assumption is made in SHRUTI, where each relation may be instantiated only a fixed number of times in one reasoning episode. This is not a general solution since even for datalogic programs — which do not need function symbols in the underlying language — exponentially many copies may be needed. The BUR system from [25] does not provide multiple copies, which is the reason for the fact that the system may be unsound if multiplace relation symbols are involved.

The results from [18] suggest that the problem of generating new instances of a rule can be mapped on the problem of obtaining a better approximation of the least model of a logic problem in the following sense: The level assigned to ground atoms occurring in the $n$-th iteration of the meaning function for the first time should be higher than the level assigned to ground atoms which occur at earlier stages. If the accuracy of an approximation can then be correlated to the available hardware resources as in [24], we might obtain a solution for this challenge problem.

### 3.6 How can insights from neuroscience be used to design integrated systems which are biologically feasible?

Artificial neural networks are very coarse abstractions of biological networks. Connectionist networks used for the study of neural-symbolic integration, however, are often biologically much less feasible than standard architectures like multilayer perceptrons. While it is important to study the formal relationships between first-order logic and connectionist systems, we believe that it is also important to study biological networks from the perspective of symbolic knowledge processing. Can the accumulation of electric potential within a dendritic tree be understood from a logial perspective? Can we develop methods to understand the temporal aspects of different transmission times between different neurons? Can we assign logical meaning to firing patterns of collections of neurons? Interdisciplinary efforts are required to answer these questions!

### 3.7 What is the exact relationship between neural-symbolic integration and chaos theory? Can this be exploited?

This question is prompted not only by our results reported in Section 2, but also by work by Blair et al. on the relationship between cellular automata, topological dynamics, logic programming, and other paradigms related to chaos theory [8]. The structural coincidences are striking, but research in this direction is difficult due to the fact that the related paradigms all turn out to be equally hard to study, and advances will most likely necessitate entirely new ideas for approaching these issues.

## 3.8 What does a theory for the integration of logic and connectionist systems look like?

The results achieved so far on connectionist inference systems are more or less unrelated to each other. Different logics are mapped onto different connectionist systems and very often not much effort has been spent on (i) formally showing properties of the system, (ii) formally relating the logic to the system and (iii) formally relating the various systems to each other. There are some exceptions though, eg. in [21] it was proven that the presented connectionist system really solves the unification and matching problem or in [6] we have given a rigorous logical reconstruction of the backward reasoning version of SHRUTI.

We would like to see a theory where in various layers of increased expressiveness logics, their corresponding connectionist models, their time and space complexities, their properties concerning learning and rule extraction as well as learning and rule extraction algorithms are specified. Such a theory could be developed along the lines proposed in [23], the BUR system, [10, 13], and [2, 18]: In each layer the logic would be defined by a certain class of logic programs and the corresponding connectionist systems would be recurrent neural networks.

For example, if the logic programs are propositional, then interpretations are represented locally. The units in the corresponding recurrent neural network are logical threshold units. If learning shall be applied, then the threshold units in the hidden layer must be replaced by sigmoidal ones. If the programs are datalogic programs, then the corresponding recurrent neural network must be able to bind variables to constants which can be done by using phase coding as in SHRUTI and BUR. If the logic programs are full first order, then interpretations shall be represented by vectors of real numbers and models are only approximated, etc.

The logics in such a theory shall not only be the standard monotonic ones, but we should also consider nonmonotonic ones. By the way, nonmonotonic reasoning was originally proposed as a technique for "jumping to a conclusion". Nowadays conventional nonmonotonic reasoning systems have time and space complexities which are not at all in accordance with the original goal. It may well be that connectionist techniques may help to put the research in nonmonotonic reasoning techniques back on track.

A general theory for the integration of logic and connectionist systems could also be developed for symmetric networks [28]. Pinkas has shown that the problem of finding a model for a propositional logic formula is equivalent to finding a global minimum in an energy function [38]. He has extended his results to some nonmonotonic [37] and first-order logics [36]. Again, the picture is far from being complete.

## 3.9 Can such a theory be applied in a real domain outperforming conventional approaches?

All applications of connectionist inference systems that we have seen so far are toy examples. We have to come up with applications in real domains which outperform conventional approaches. This can only be done if we use hardware which exploits the massive parallelism of connectionist networks. If we are reasoning in a logic whose entailment problem is in $\mathcal{NC}$ and an efficient or optimal parallel algorithm for deciding this problem is known, then it does not suffice to simulate this algorithm on a computer with just a few processors.

Because a general theory for integrating logic and connectionist systems may be layered, applications we are looking for should have a similar structure. It may be worth while to look for such an application in the area of integrating the low-level control of a real robot with the high-level control developed in the area of cognitive robotics. Such an application would also be a good showcase for learning and rule extraction: Even if such a robot is initialized with some knowledge, it must learn to behave in its environment, and this learning never stops. Consequently, the knowledge is constantly updated.

## 4  Summary

In this paper we have given an overview on how first-order logic programs can be represented in a connectionist setting and outlined various challenges for developing a truly connectionist system capable of representing structured objects and performing structure-sensitive processes.